\documentclass{article}

\usepackage[final]{corl_2018}

\usepackage[utf8]{inputenc} 
\usepackage[T1]{fontenc}    
\usepackage{hyperref}       
\usepackage{url}            
\usepackage{booktabs}       
\usepackage{amsfonts}       
\usepackage{nicefrac}       
\usepackage{microtype}      

\usepackage{amsmath,amsfonts,amssymb}
\usepackage{algorithm,algorithmicx}
\usepackage[noend]{algpseudocode}
\usepackage{graphicx}
\usepackage{caption}
\usepackage{subcaption}
\usepackage{mathtools}
\usepackage{tikz}
\usepackage{pgf}
\usetikzlibrary{arrows,automata}
\usepackage{multicol}

\newcommand{\abf}{\mathbf{a}}

\newcommand{\fbf}{\mathbf{f}}
\newcommand{\gbf}{\mathbf{g}}

\newcommand{\kbf}{\mathbf{k}}

\newcommand{\qbf}{\mathbf{q}}
\newcommand{\rbf}{\mathbf{r}}
\newcommand{\sbf}{\mathbf{s}}

\newcommand{\ubf}{\mathbf{u}}

\newcommand{\xbf}{\mathbf{x}}
\newcommand{\ybf}{\mathbf{y}}
\newcommand{\zbf}{\mathbf{z}}

\newcommand{\Abf}{\mathbf{A}}

\newcommand{\Ebf}{\mathbf{E}}

\newcommand{\Hbf}{\mathbf{H}}
\newcommand{\Ibf}{\mathbf{I}}
\newcommand{\Jbf}{\mathbf{J}}
\newcommand{\Kbf}{\mathbf{K}}
\newcommand{\Lbf}{\mathbf{L}}
\newcommand{\Mbf}{\mathbf{M}}

\newcommand{\Qbf}{\mathbf{Q}}

\newcommand{\Sbf}{\mathbf{S}}

\newcommand{\Xbf}{\mathbf{X}}

\newcommand{\Zbf}{\mathbf{Z}}

\newcommand{\Acal}{\mathcal{A}}

\newcommand{\Lcal}{\mathcal{L}}

\newcommand{\Ncal}{\mathcal{N}}
\newcommand{\Ocal}{\mathcal{O}}

\newcommand{\Scal}{\mathcal{S}}

\newcommand{\Xcal}{\mathcal{X}}

\newcommand{\alphabf}{\boldsymbol{\alpha}}

\newcommand{\thetabf}{\boldsymbol{\theta}}
\newcommand{\Thetabf}{\mathbf{\Theta}}
\newcommand{\Lambdabf}{\mathbf{\Lambda}}

\newcommand{\varepsilonbf}{\boldsymbol{\varepsilon}}

\newcommand{\tr}{\text{tr}}
\newcommand{\zerobf}{\mathbf{0}}
\newcommand{\ellbf}{\boldsymbol{\ell}}

\newcommand{\etal}{\textit{et al. }}

\DeclareMathOperator*{\argmax}{arg\,max}

\title{Sparse Gaussian Process Temporal Difference Learning for Marine Robot Navigation}

\author{
  John Martin, Jinkun Wang, Brendan Englot\\
  Department of Mechanical Engineering\\
  Stevens Institute of Technology\\
  \texttt{\{jmarti3, jwang92, benglot\}@stevens.edu} \\
}

\begin{document}
\maketitle

%
%
\begin{abstract}
We present a method for Temporal Difference (\textsc{td}) learning that addresses several challenges faced by robots learning to navigate in a marine environment. For improved data efficiency, our method reduces \textsc{td} updates to Gaussian Process regression. To make predictions amenable to online settings, we introduce a sparse approximation with improved quality over current rejection-based methods. We derive the predictive value function posterior and use the moments to obtain a new algorithm for model-free policy evaluation, \textsc{spgp-sarsa}. With simple changes, we show \textsc{spgp-sarsa} can be reduced to a model-based equivalent, \textsc{spgp-td}. We perform comprehensive simulation studies and also conduct physical learning trials with an underwater robot. Our results show \textsc{spgp-sarsa} can outperform the state-of-the-art sparse method, replicate the prediction quality of its exact counterpart, and be applied to solve underwater navigation tasks.	
\end{abstract}

%
%

%
%
\section{Introduction}

Model-free Reinforcement Learning (\textsc{rl}) demands that robots produce lots of data to evaluate and improve their decision making policies. For marine robots, this can be challenging, since learning must be performed online, and their acoustics-based sensors produce data in low volumes. Here, we will present an algorithm that supports the learning of navigation policies with very little data.

Our algorithm belongs to the class of value estimation methods. Such methods are rooted in Bellman's equation, which describes the value of taking action $\abf$ in state $\sbf$ as a sum of the expected reward and the forecasted value over a random transition $(\sbf,\abf) \rightarrow (\Sbf,\Abf)$: 
\begin{align*}
	Q(\sbf, \abf) = \Ebf[R(\sbf,\abf)] + \gamma \Ebf[Q(\Sbf,\Abf)]. 
\end{align*}
The contractive nature of Bellman's equation motivates many standard algorithms, which apply the recursion to obtain better estimates of the true value \cite{rl:2010:szepesvari:algorithms_for_reinforcement_learning,rl:1988:sutton:learning_to_predict_by_the_methods_of_temporal_differences, rl:1994:rummery_niranjan:online_q_learning_using_connectionist_systems}.
Our method chooses a less conventional approach, but one which is more data-efficient, reducing Bellman updates to Gaussian Process (\textsc{gp}) regression \cite{rl:2003:engel_mannor_meir:bayes_meets_bellman_the_gaussian_process_approach_to_temporal_difference_learning}. 

\textsc{gp} regression is a flexible Bayesian method for learning unknown functions from data \cite{gp:rasmussen_williams-gaussian_processs_for_machine_learning}. It uses a kernel-based covariance structure to promote a high degree of data association, well suited for learning when data is scarce. The main drawback is the prediction complexity, which scales prohibitively as $\Ocal(N^3)$, where $N$ is the number of data points \cite{gp:rasmussen_williams-gaussian_processs_for_machine_learning}. To address this issue, we appeal to a sparse approximation.

Sparse approximations have been proposed to reduce the complexity of exact predictions \cite{gp:2002:csato_sparseon:sparse_online_gaussian_processes,rl:2003:engel_mannor_meir:bayes_meets_bellman_the_gaussian_process_approach_to_temporal_difference_learning, rl:2005:engel_mannor_meir:reinforcement_learning_with_gaussian_processes,rl:2011:jakab_csato:improving_gaussian_process_value_function_approximation_in_policy_gradient_algorithms}. Many methods use an information criterion as the basis for rejecting redundant data. For \textsc{gp-rl}, Engel \etal employ the conditional covariance as a measure of relative error \cite{rl:2003:engel_mannor_meir:bayes_meets_bellman_the_gaussian_process_approach_to_temporal_difference_learning,rl:2005:engel_mannor_meir:reinforcement_learning_with_gaussian_processes}, and they discard points that contribute little error to predictions. By limiting the active set to $M \ll N$ points, their predictions never cost more than $NM^2$ operations. 

%

Besides discarding potentially useful information, the most prevalent issue with rejection sparsification is how it interferes with model selection. It can cause sharp changes to appear in the marginal likelihood and complicate evidence maximization with common optimizers, such as \textsc{l-bfgs} \cite{rl:1989:liu_nocedal:limited_memory_bfgs}. 

In what follows, we will describe a new method for \textsc{gp} value estimation that induces sparsity without discarding data. We approximate the exact \textsc{td} regression framework of Engel \etal \cite{rl:2003:engel_mannor_meir:bayes_meets_bellman_the_gaussian_process_approach_to_temporal_difference_learning} with a smaller active set containing $M$ adjustable support points. With this change, our method achieves the same prediction complexity as the state-of-the-art approximation, $\Ocal(NM^2)$, while incurring less approximation error. Our method is based on the Sparse Pseudo-input Gaussian Process (\textsc{spgp}) \cite{gp:2006:snelson_ghahramani:sparse_gaussian_processes_using_pseudo_inputs}. Therefore, it inherits many of \textsc{spgp}'s favorable characteristics, all of which we show support the unique needs of robots learning to navigate in a marine environment. 

\section{TD Value Estimation as GP Regression}\label{sec:gptd}
\textsc{td} algorithms recover the latent value function with data gathered in the standard \textsc{rl} fashion: at each step, the robot selects an action $\abf\in\Acal$ based on its current state $\sbf\in\Scal$, after which it transitions to the next state $\sbf'$ and collects a reward $R\sim p_r(\cdot|\sbf,\abf)$. The repeated interaction is described as a Markov Decision Process, $(\Scal,\Acal,p_r,p_s,\gamma)$, associated with the transition distribution $\sbf'\sim p_s(\cdot|\sbf,\abf)$, stationary policy $\abf\sim\pi(\cdot|\sbf)$, and discount factor $\gamma \in [0,1]$. As the name suggests, \textsc{td} algorithms update a running estimate of the value function to minimize its difference from the Bellman estimate: $r + \gamma Q(\sbf',\abf') - Q(\sbf, \abf)$; $r$ being the observed reward. Once the estimate converges, a robot can navigate by selecting actions from the greedy policy $\pi$, such that $\Ebf_\pi[\Abf|\sbf] = \argmax_{\abf\in \Acal} Q(\sbf,\abf)$.

The Gaussian Process Temporal Difference (\textsc{gptd}) framework improves upon the data efficiency of frequentist \textsc{td} estimation by departing from the contractive nature of Bellman's equation, in favor of a convergence driven by non-parametric Bayesian regression. The data model is based on the random return $Z(\xbf) = \sum_{t=0}^\infty \gamma^tR(\xbf_t)$, expressed as a sum of its mean, $Q(\xbf)$, and zero-mean residual, $\Delta Q(\xbf) = Z(\xbf) - Q(\xbf)$. Model inputs are state-action vectors $\xbf\in \Xcal=\Scal\times\Acal$, and value differences are used to describe the observation process:
\begin{align}\label{eq:trans_likelihood}
	R(\xbf) &= Q(\xbf) -\gamma Q(\xbf') + [\Delta Q(\xbf)-\gamma\Delta Q(\xbf')] = Q(\xbf) -\gamma Q(\xbf') + \varepsilon(\xbf,\xbf').
\end{align}
Our work makes the simplifying assumption that noise levels, $\varepsilon(\xbf,\xbf')$, are i.i.d random variables with constant parameters, $\varepsilon \sim \Ncal(0,\sigma^2)$. Under this assumption, transitions exhibit no serial correlation. Although it is straightforward to model serially-correlated noise \cite{rl:2005:engel_mannor_meir:reinforcement_learning_with_gaussian_processes}, doing so would preclude application of our desired approximation. We elaborate on this point in Section \ref{sec:sgptd}.


Given a time-indexed sequence of transitions $(\xbf_{t},R(\xbf_{t}),\xbf_{t+1})_{t=0}^{N-1}$, we stack the model variables into vectors, resulting in the complete data model: $\rbf = \Hbf\qbf(\xbf)+\varepsilonbf$, where  $\qbf \sim \Ncal(\zerobf,\Kbf_{qq})$, and
\begin{align}\label{eq:gptd_model}
	\begin{pmatrix}
		R(\xbf_0)\\
		R(\xbf_1)\\
		\vdots\\
		R(\xbf_{N-1})
	\end{pmatrix} &= 
	\begin{pmatrix}
		1 & -\gamma & 0 & \cdots & 0\\
		0 & 1 & -\gamma & \cdots &0 \\
		\vdots & & & & \vdots\\
		0 & 0 &\cdots & 1 & -\gamma
	\end{pmatrix}
	\begin{pmatrix}
		Q(\xbf_0)\\
		Q(\xbf_1)\\
		\vdots\\
		Q(\xbf_N)
	\end{pmatrix} + 
	\begin{pmatrix}
		\varepsilon_0\\
		\varepsilon_1\\
		\vdots\\
		\varepsilon_N
	\end{pmatrix}.
\end{align}
Notice the commonality Equation \ref{eq:gptd_model} has with a standard \textsc{gp} likelihood model, $\ybf = \fbf(\xbf) + \varepsilonbf$. Both models assume the outputs, $\rbf \sim \ybf$, are noisy observations of a latent function, $\qbf \sim \fbf$. What distingushes \textsc{td} estimation is the presence of value correlations, imposed from Bellman's equation and encoded as temporal difference coefficients in $\Hbf$. Used for exact \textsc{gp} regression, Equation \ref{eq:gptd_model} leads to the \textsc{gp-sarsa} algorithm: a non-parametric Bayesian method for recovering latent values \cite{rl:2003:engel_mannor_meir:bayes_meets_bellman_the_gaussian_process_approach_to_temporal_difference_learning}. 

As a Bayesian method, \textsc{gp-sarsa} computes a predictive posterior over the latent values by conditioning on observed rewards. The corresponding mean and variance are used for policy evaluation:
\begin{align}
	v(\xbf_*) &= \kbf^\top_{r*}(\Kbf_{rr} + \sigma^2\Ibf)^{-1}\rbf, &
	s(\xbf_*) &= k(\xbf_*,\xbf_*) -\kbf^\top_{r*}(\Kbf_{rr} + \sigma^2\Ibf)^{-1}\kbf_{r*}.
	\label{eq:gptd_predmoms}
\end{align}
Here, $\Kbf_{qq}$ is the covariance matrix with elements  $[\Kbf_{qq}]_{ij} = k(\xbf_i,\xbf_j)$, $\Kbf_{rr} = \Hbf\Kbf_{qq}\Hbf^\top$, and $\kbf_{r*} = \Hbf\kbf_*$, where $[\kbf_*]_{i} = k(\xbf_i,\xbf_*)$. Subscripts denote dimensionality, e.g. $\Kbf_{qq} \in \mathbb{R}^{|\qbf|\times |\qbf|}$.

%
%
\section{Sparse Pseudo-input Gaussian Process Temporal Difference Learning} \label{sec:sgptd}
\begin{figure*}[]
	\centering
	\includegraphics[width=0.31\textwidth]{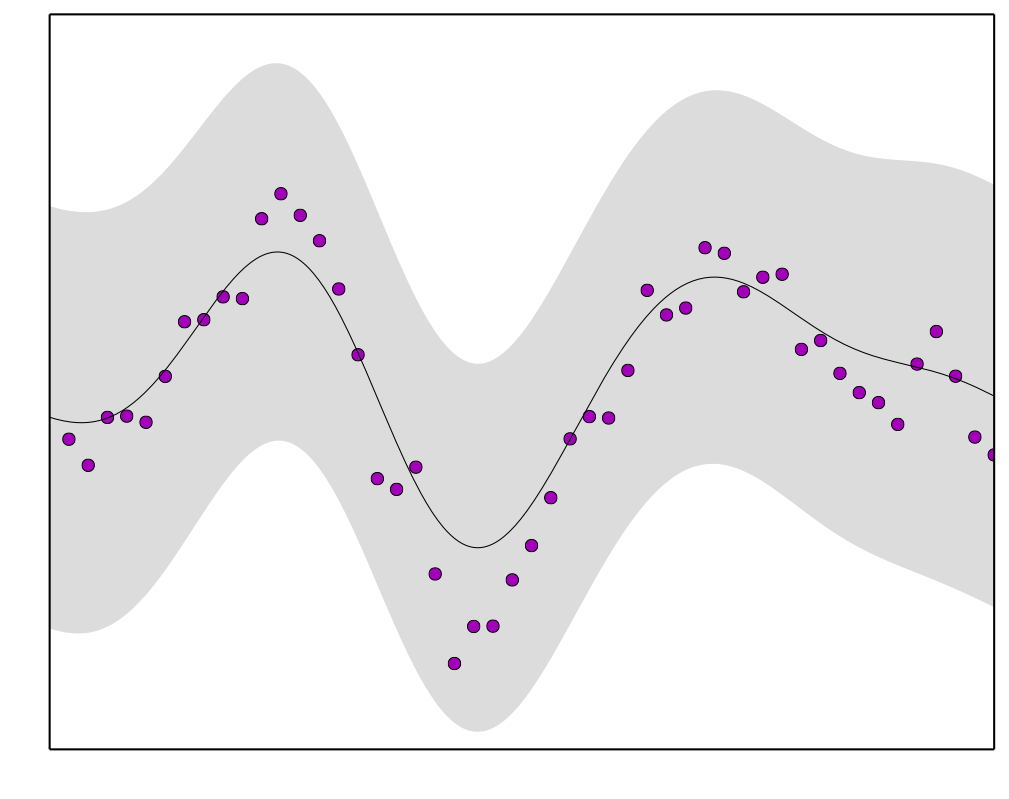}
	\includegraphics[width=0.31\textwidth]{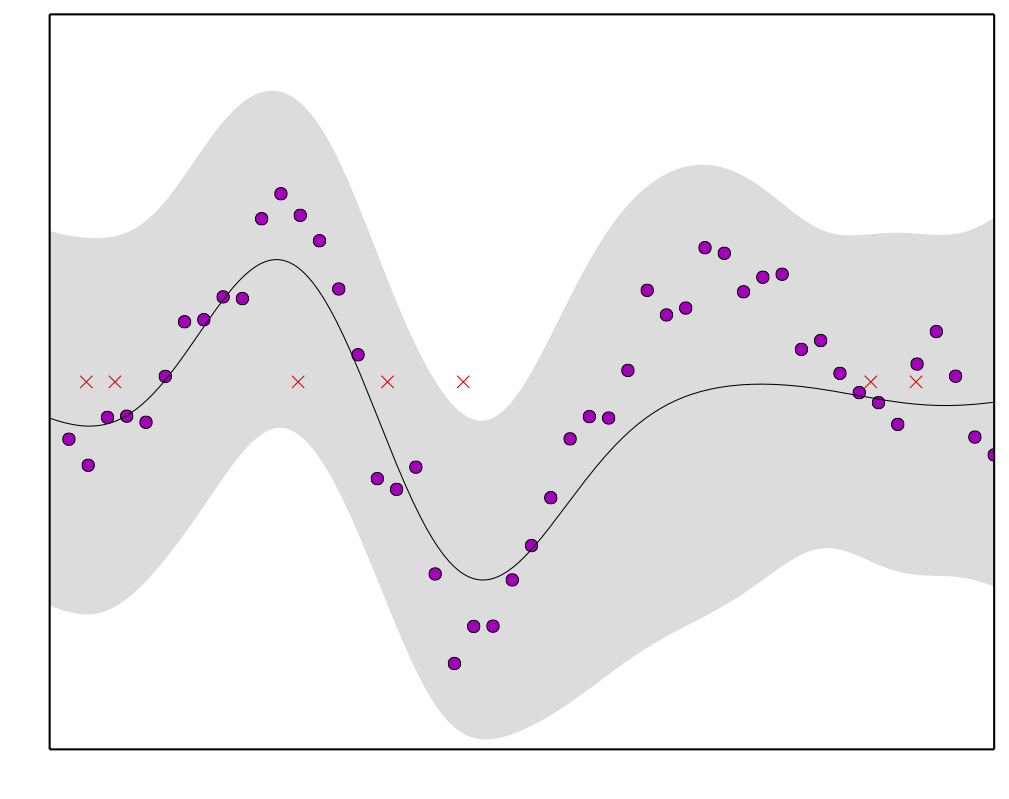}
	\includegraphics[width=0.31\textwidth]{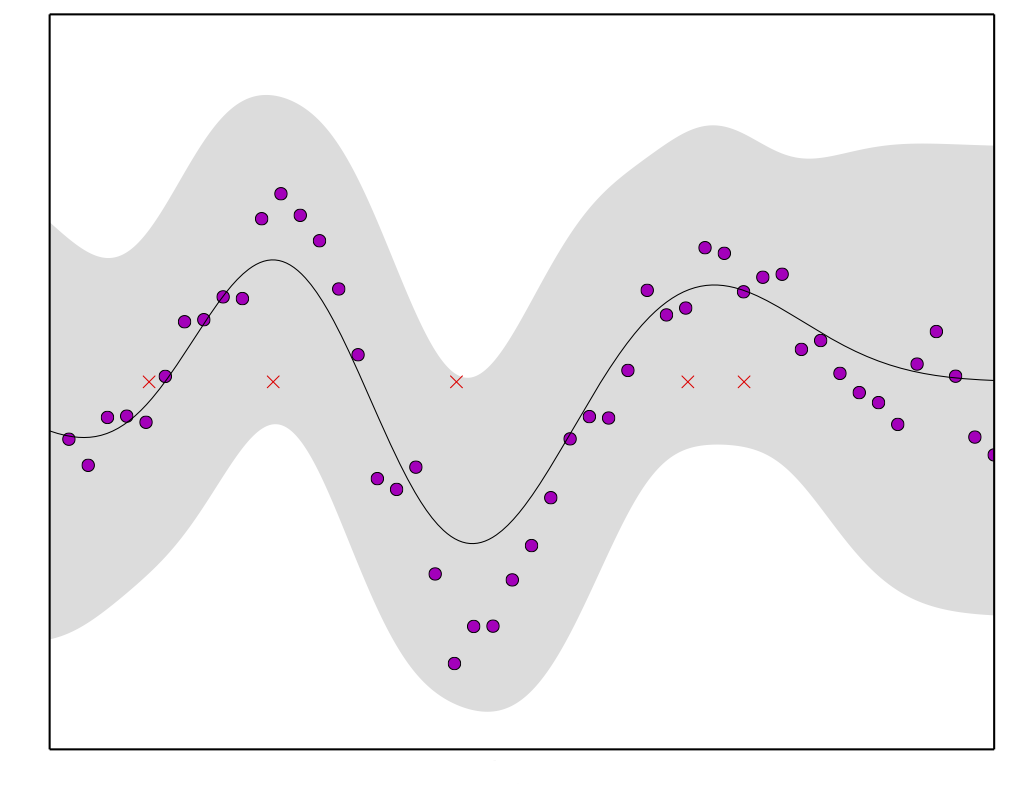}
	\caption{\textbf{Visualizing the approximate posterior:} We plot the predictive distributions of \textsc{gp-sarsa} (left) and our approximate method, \textsc{spgp-sarsa}. We used $M = 7 $ randomly-initialized pseudo inputs (red crosses) and $N=50$ training samples (magenta) taken from the prior. The predictive mean is black, and two standard deviations of uncertainty are shown in gray. We plot \textsc{spgp-sarsa} before (center) and after pseudo input optimization (right). After training, the sparse method is nearly identical to the exact method.}\label{fig:posterior_samples}
	\vspace{-1em}
\end{figure*}

The \textsc{gp-sarsa} method requires an expensive $N\times N$ matrix inversion, costing $\Ocal(N^3)$. Since robots must estimate values online to improve their navigation policies, we appeal to an approximate method which induces sparsity in the standard data model (Equation \ref{eq:gptd_model}). We expand the probability space with $M\ll N$ additional pseudo values, $\ubf$. Here, pseudo values serve a parametric role in a non-parametric setting; they are free variables that provide probability mass at support locations $\zbf\in\Zbf\subset \Xcal$. The extra latent variables obey the same data model as $\qbf$, but are predetermined, and thus, exhibit no noise. By conditioning $\qbf$ upon $\ubf$ and $\Zbf$, we can collapse the predictive probability space such that all dense matrix inversions are of rank $M$. Finally, we optimize $\Zbf$ to maximize the likelihood and produce a high-quality fit. This strategy is called Sparse Pseudo-input \textsc{gp} regression \cite{gp:2006:snelson_ghahramani:sparse_gaussian_processes_using_pseudo_inputs} (Figure \ref{fig:posterior_samples}).

Although \textsc{spgp} regression is well-known, it has never been applied to \textsc{td} estimation, where latent variables exhibit serial correlation. Therefore, current results from the sparse \textsc{gp} literature do not apply. In what follows we apply the \textsc{spgp} principles to derive a new method for \textsc{td} value estimation, which we call \textsc{spgp-sarsa}. The method is data-efficient and fast enough for online prediction.

\subsection{Latent Value Likelihood Model}
 Given an input location, $\xbf$, the likelihood of the latent value, $Q(\xbf)$, is the conditional probability
\begin{align}
	p(Q|\xbf,\Zbf,\ubf) &= \Ncal(Q | \kbf_{u}^\top\Kbf_{uu}^{-1}\ubf, k(\xbf,\xbf) - \kbf_{u}^\top\Kbf_{uu}^{-1}\kbf_{u}),
\end{align}
where $[\Kbf_{uu}]_{ij} = k(\zbf_i,\zbf_j)$, $[\kbf_u]_i = k(\zbf_i,\xbf)$. The complete data likelihood is obtained by stacking the $N$ independent single transition likelihoods
\begin{align}
	p(\qbf | \Xbf,\Zbf,\ubf) &= \prod_{t=1}^Np(Q_t|\xbf_t,\Zbf,\ubf) = \Ncal( \gbf, \widetilde{\Kbf});
\end{align}
where we defined $\gbf = \Kbf_{qu}\Kbf_{uu}^{-1}\ubf$, $\widetilde{\Kbf} = \text{diag}(\Kbf_{qq} - \Kbf_{qu}\Kbf_{uu}^{-1}\Kbf_{uq})$, with $[\Kbf_{qu}]_{ij} = k(\xbf_i,\zbf_j)$. 

\subsection{Conditioned Data Likelihood Model}
Here we derive the likelihood distribution of the observed rewards conditioned on the pseudo values $p(\rbf|\Xbf,\Zbf,\ubf)$. Consider the transformed joint distribution over values, rewards, and pseudo values 
\begin{align*}
	\begin{pmatrix}
		\qbf\\
		\rbf\\
		\ubf
	\end{pmatrix} \sim
	\Ncal\left(
	\begin{pmatrix}
		\zerobf\\
		\zerobf\\
		\zerobf
	\end{pmatrix},
	\begin{pmatrix}
		\Kbf_{qq} & \Kbf_{qq}\Hbf^\top & \Kbf_{qu}\\
		\Hbf\Kbf_{qq} & \Hbf\Kbf_{qq}\Hbf^\top+\sigma^2\Ibf & \Hbf\Kbf_{uq}\\
		\Kbf_{uq} & \Kbf_{uq}\Hbf^\top & \Kbf_{uu}
	\end{pmatrix}
	\right).
\end{align*}
The likelihood is obtained by conditioning $\rbf$ on $\ubf$ and invoking transition independence:
\begin{align}
	p(\rbf|\Xbf,\Zbf,\ubf) &= \Ncal(\Kbf_{ru}\Kbf_{uu}^{-1}\ubf,\Qbf + \sigma^2\Ibf),\label{eq:pru}&
	\Qbf &= \text{diag}(\Kbf_{rr} - \Kbf_{ru}\Kbf_{uu}^{-1}\Kbf_{ur}).
\end{align}

\subsection{Posterior of Pseudo Values}
To obtain the posterior $p(\ubf|\rbf,\Xbf,\Zbf)$, we use Bayes' rule. Given the marginal $p(\ubf|\Zbf) = \Ncal(\ubf|\mathbf{0}, \Kbf_{uu})$ and the conditional for $\rbf$ given $\ubf$ (Equation \ref{eq:pru}), the posterior for $\ubf$ given $\rbf$ is
\begin{align}
	p(\ubf|\rbf,\Xbf,\Zbf) &= \Ncal(\ubf|\Lbf\Kbf_{uu}^{-1}\Kbf_{ur}(\Qbf + \sigma^2\Ibf)^{-1}\rbf,\Lbf);\\
	\Lbf&= \Kbf_{uu}(\Kbf_{uu} + \Kbf_{ur}(\Qbf + \sigma^2\Ibf)^{-1}\Kbf_{ru})^{-1}\Kbf_{uu}.\nonumber
\end{align}

\subsection{Latent Value Predictive Posterior}
The predictive posterior is obtained by marginalizing the pseudo values:
\begin{align*}
	p(Q_*|\xbf_{*},\rbf,\Xbf,\Zbf) &= \int p(Q_* | \xbf_*,\Zbf,\ubf)p(\ubf|\rbf,\Xbf,\Zbf) d\ubf.
\end{align*}
Let $\Mbf = \Kbf_{uu} + \Kbf_{ur}(\Qbf + \sigma^2\Ibf)^{-1}\Kbf_{ru}$. Our new predictive is Gaussian with mean and variance
\begin{align}\label{eq:predictive_mean}
	\tilde{v}(\xbf_*) &=\kbf_{u*}^\top\underbrace{\Mbf^{-1}\Kbf_{ur}(\Qbf + \sigma^2\Ibf)^{-1}\rbf}_{\alphabf_\pi},&
	\tilde{s}(\xbf_*) &= k(\xbf_*,\xbf_*) - \kbf_{u*}^\top\underbrace{(\Kbf_{uu}^{-1} - \Mbf^{-1})}_{\Lambdabf_\pi}\kbf_{u*}.
\end{align}
Equation \ref{eq:predictive_mean} represents our main contribution. The parameters $\alphabf_\pi$ and $\Lambdabf_\pi$ are independent of the input, and their expressions depend on two inverses. The first is the dense $M$-rank matrix, $\Kbf_{uu}$. The second is the $N$-rank diagonal matrix $(\Qbf + \sigma^2\Ibf)$. When $M\ll N$, both matrices are easier to invert than a dense $N$-rank matrix. Thus, Equation \ref{eq:predictive_mean} provides motivation for estimating and predicting latent values efficiently.

\subsection{Assumptions and Related Work }
There are several key assumptions underpinning our results. To guarantee the likelihood can be factored, we need to assume that transitions are uncorrelated. Had we modeled serial correlation in the noise process, $\varepsilonbf$ would be distributed with a tridiagonal covariance \cite{rl:2005:engel_mannor_meir:reinforcement_learning_with_gaussian_processes}, which cannot be factored directly. It is possible to obtain a factorable model by applying a whitening transform. However, this changes the observation process to Monte Carlo samples, which are known to be noisy \cite{rl:1998:sutton_barto:introduction_to_reinforcement_learning}. Under our simpler set of assumptions, we obtain an efficiently-computable, sparse representation of the value posterior that is amenable to smooth evidence maximization. Section \ref{sec:finding_hypers_and_pseudos} details how to select the hyperparameters and pseudo inputs with gradient-based optimization.

The most relevant methods to our work apply \textsc{gp} regression to estimate the latent value function in a \textsc{td} setting \cite{rl:2003:engel_mannor_meir:bayes_meets_bellman_the_gaussian_process_approach_to_temporal_difference_learning, rl:2005:engel_mannor_meir:reinforcement_learning_with_gaussian_processes,rl:2005:engel:algorithms_and_representations_for_reinforcement_learning,rl:2005:engel_etal:learning_to_control_an_octopus_arm_with_gaussian_process_temporal_difference_methods}. This class of algorithms is distinct from those which apply \textsc{gp} regression in the absence of sequential correlation, with Monte Carlo sample returns \cite{rl:2010:deisenroth:efficient_reinforcement_learning_using_gaussian_processes:thesis}, and methods whose convergence behavior is driven primarily by the Bellman contraction \cite{rl:2009:deisenroth_rasmussen_peters:gaussian_process_dynamic_programming,rl:2004:rasmussen_kuss:gaussian_processes_in_reinforcement_learning}. As a \textsc{gp-td} algorithm, the policy update process (Algorithm \ref{alg:policyiter}) depends only on \textsc{gp} regression. This convergence is known to be data efficient and asymptotically unbiased \cite{gp:rasmussen_williams-gaussian_processs_for_machine_learning}. We do not prove convergence for \textsc{gp-td} algorithms here; however, we mention the convergence behavior to underscore our method's relevance to robots learning with limited volumes of data. We also note these methods have been scaled to high-dimensional systems with complex, continously-varying dynamics \cite{rl:2005:engel_etal:learning_to_control_an_octopus_arm_with_gaussian_process_temporal_difference_methods}.

When it comes to other approximate \textsc{gp-td} methods, the state-of-the-art uses a low-rank approximation to the full covariance matrix \cite{rl:2005:engel_mannor_meir:reinforcement_learning_with_gaussian_processes}: $\Kbf_{qq}\approx \Abf\tilde{\Kbf}^{-1}\Abf^\top$, where $\Abf$ is a projection matrix. Before adding new data to the active set, \textsc{lowrank-sarsa} checks if new data points increase the conditional covariance by a desired error threshold, $\nu$. In Section \ref{sec:approx_quality}, we compare the approximation quality of \textsc{lowrank-sarsa} and \textsc{spgp-sarsa}.

%
%
\section{New Algorithms for Robot Navigation}
\begin{multicols}{2}
	\begin{algorithm}[H]
		\caption{\textsc{spgp-sarsa}}
		\label{alg:spgpsarsa}
		\begin{algorithmic}[1]
		\State \textbf{input: $\xbf_0$, $\pi$, $\thetabf$, $\Zbf$} 
			\For{$t=1,\cdots,N$}
				\State \text{Observe } $\xbf_{t-1}, r_{t-1}, \xbf_t$
			\EndFor
			\State $\alphabf_\pi \gets\Mbf^{-1}\Kbf_{ur}(\Qbf + \sigma^2\Ibf)^{-1}\rbf$
			\State $\Lambdabf_\pi \gets\Kbf_{uu}^{-1} - \Mbf^{-1}$	
			\State Maximize Eq. \ref{eq:likelihood} for optimal $\thetabf$, $\Zbf$ (Optional)
			\State \textbf{output:} $\alphabf_\pi,\Lambdabf_\pi$
		\end{algorithmic}
	\end{algorithm}
	
	\columnbreak
	
	\begin{algorithm}[H]
		\caption{\textsc{policy-iteration}}
		\label{alg:policyiter}
		\begin{algorithmic}[1]
		\State \textbf{input: $\xbf_0$, $\thetabf$, $\Zbf$} 
		\State $\pi\gets \pi_{\text{init}}$
			\State \textbf{repeat:} 
				\State \indent $\alphabf_\pi,\Lambdabf_\pi\gets$\textsc{spgp-sarsa}$(\xbf_0,\pi,\thetabf,\Zbf)$
				\State \indent $\pi\gets$\textsc{greedy-update}$(\alphabf_\pi,\Lambdabf_\pi)$
			\State \textbf{until:} $\pi$ converges 
			\State \textbf{output:} $\pi$
		\end{algorithmic}
	\end{algorithm}	
\end{multicols}

Navigation tasks are specified through the reward function. As a robot transitions through its operating space, it should assign the highest value to states and actions that bring it closer to the goal, and the lowest values around obstacles and other forbidden regions. We provide examples of such functions in Section \ref{sec:results}. 

Given a suitable reward function, Algorithm \ref{alg:spgpsarsa} implements \textsc{spgp-sarsa} regression, where the posterior value function parameters $\alphabf_\pi,\Lambdabf_\pi$ are learned with sequentially-observed data. The policy is updated using standard policy iteration, described in Algorithm \ref{alg:policyiter}.

\subsection{Finding the Hyper-parameters and Pseudo-inputs}\label{sec:finding_hypers_and_pseudos}
\begin{figure}
	\centering
	\begin{subfigure}[t]{0.45\textwidth}
		\includegraphics[width=\columnwidth]{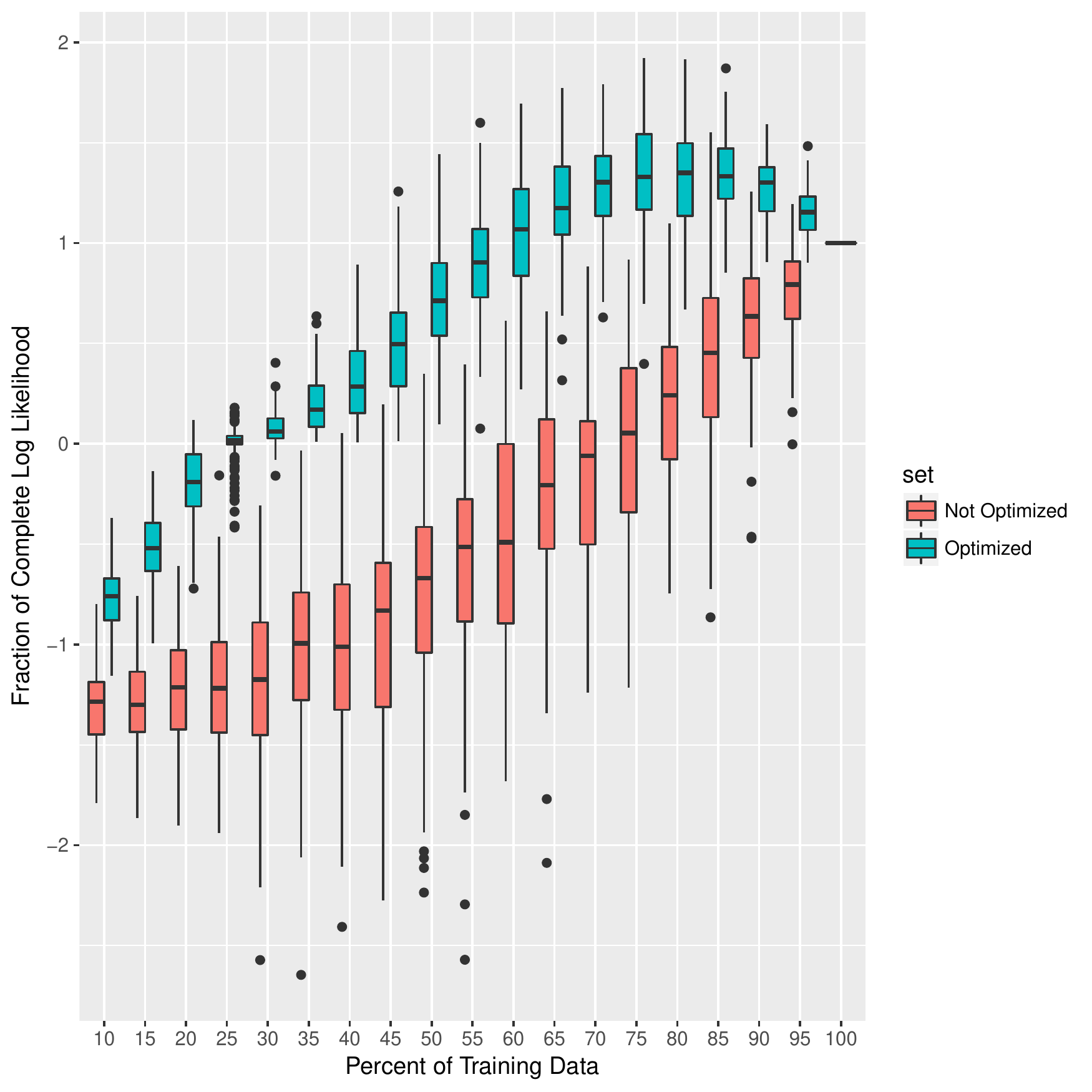}
		\caption{\textsc{spgp-sarsa} likelihood disparity}\label{fig:sparse_tests:likedisp}
	\end{subfigure}
	\begin{subfigure}[t]{0.45\textwidth}
		\centering
		\includegraphics[width=0.55\columnwidth]{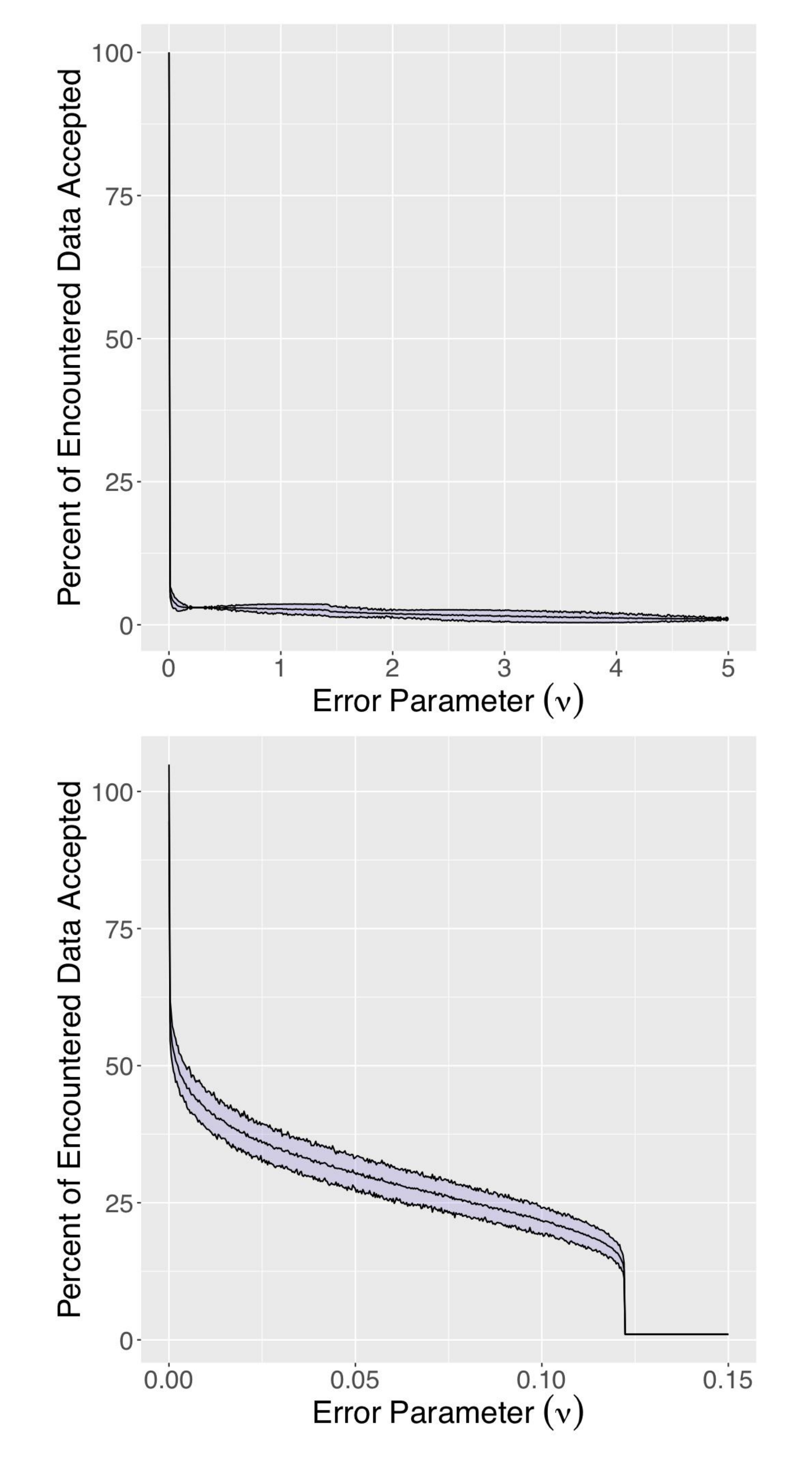}
		\caption{\textsc{lowrank-sarsa} rejection statistics}\label{fig:sparse_tests:rejstats}
	\end{subfigure}
	\caption{\textbf{Visualizing the effect of parameter changes on approximation quality:} Figure \ref{fig:sparse_tests:likedisp} shows how the inclusion of more pseudo-inputs improves approximation quality, and how optimizating their locations is additionally benificial. We fixed a synthetic dataset of 100 samples from the prior, $\Ncal(\zerobf,\Kbf_{qq})$. As a reflection of quality, we uniformly sampled subsets of the data, computed the log likelihood with \textsc{spgp-sarsa}, both before and after optimization, and normalized it by the log likelihood of the full set. Boxplots were computed for 100 random subsets. Figure \ref{fig:sparse_tests:rejstats} shows how the low-rank method cannot always induce sparsity. We vary the error threshold, $\nu$, and plot the percentage of samples \textsc{lowrank-sarsa} retained from 100 random trajectories. Trajectories from the Mountain Car system (top), and prior,  $\Ncal(\zerobf,\Kbf_{qq})$ (bottom) are shown.}
	\label{fig:sparse_tests}
	\vspace{-1em}
\end{figure}

We use the marginal likelihood to fit the hyper-parameters $\Thetabf=\{\thetabf, \sigma^2\}$ and pseudo inputs $\Zbf$ to the observed data, $\rbf,\Xbf$. Unlike rejection-based sparsification, our method has the benefit of being continuous in nature. This allows for the variables to be tuned precisely to achieve a high-quality fit.  The marginal likelihood is a Gaussian, given by
\begin{align}\label{eq:likelihood}
	p(\rbf |\Xbf , \Zbf ) &= \int p(\rbf|\Xbf,\Zbf,\ubf)p(\ubf|\Zbf) d\ubf
	= \Ncal( \rbf | \zerobf , \Qbf + \sigma^2\Ibf + \Kbf_{ru}\Kbf_{uu}^{-1}\Kbf_{ur}).
\end{align}
Instead of optimizing Equation \ref{eq:likelihood} directly, we maximize its logarithm, $\Lcal = \log p(\rbf|\Xbf,\Zbf)$.  Given the gradient of $\Lcal$, we can use an iterative method, such as \textsc{l-bfgs}, to find the optimal parameters. Full details of the gradient computation are provided in the supplement.   

\subsection{Training Considerations} 
The frequency with which model parameters are optimized can greatly influence the runtime of Algorithm \ref{alg:spgpsarsa}. For $D$-dimensional inputs, \textsc{spgp-sarsa} must fit $DM + |\Thetabf|$ variables; whereas, \textsc{gp-sarsa} must fit only $|\Thetabf|$. Although it is not strictly necessary to refit  model parameters at each time step, the frequency which updates are needed will depend on how well $\Xbf$ and $\Zbf$ reflect the support of the operating space. As new regions are explored, the model will need to be refit. 

Even with a strategy to fit model parameters, we must still choose the number of pseudo inputs, $M$. In Figure \ref{fig:sparse_tests}, we examine the tradeoff between efficiency and accuracy in relation to $M$. As $M$ increases, $\Zbf$ begins to coincide with $\Xbf$, and prediction efficiency reduces, but the approximation improves to match the exact predictive posterior. 

Another consideration, when training \textsc{gp} models, is preventing overfitting. Overfitting occurs when the predictive variance collapses around the training data. It can be prevented by adding a regularization term to the log of Equation \ref{eq:likelihood}, penalizing the magnitude of covariance parameters and pseudo inputs.

\subsection{SPGP TD Learning}
To reduce \textsc{spgp-sarsa} to its model-based equivalent, we let $\Xcal=\Scal$ and swap the state-action transition process for the associated state transition process (Table \ref{tab:alg_map}); the analysis from Section \ref{sec:gptd} and \ref{sec:sgptd} follows directly. The new input variable, $\xbf=\sbf$, simply reduces the latent function space to one over $\text{dim}(\sbf)$ variables. Equation \ref{eq:predictive_mean} describes the state value posterior moments, analogous to frequentist \textsc{td} \cite{rl:1998:sutton_barto:introduction_to_reinforcement_learning} and standard \textsc{gp-td} \cite{rl:2003:engel_mannor_meir:bayes_meets_bellman_the_gaussian_process_approach_to_temporal_difference_learning,rl:2005:engel:algorithms_and_representations_for_reinforcement_learning}. We call the resulting algorithm \textsc{spgp-td}.
\begin{table}[H]
	\centering
	\begin{tabular}{r|c|c}
		Target value $\tilde{v}(\xbf)$ & Input space $\Xcal$ & Transition Dist. $p_\pi(\xbf_t|\xbf_{t-1})$ \\
		\hline
		$V(\sbf)$ & $\Scal$ & $\int_{\Acal}p(\sbf_t|\sbf_{t-1},\abf_{t-1})\pi(\abf_{t-1}|\sbf_{t-1}) d\abf$\\
		$Q(\sbf,\abf)$ & $\Scal\times \Acal$ & $p(\sbf_t|\sbf_{t-1},\abf_{t-1})\pi(\abf_t|\sbf_t)$
		\vspace{0.5em}
	\end{tabular}
	
	\caption{\textbf{Mapping from SARSA to TD:} Our method can estimate values under two processes.}\label{tab:alg_map} 
	\vspace{-2em}
\end{table}

%
%
\section{Experimental Results}\label{sec:results}
\begin{figure*}
	\centering
		\includegraphics[width=0.3\textwidth]{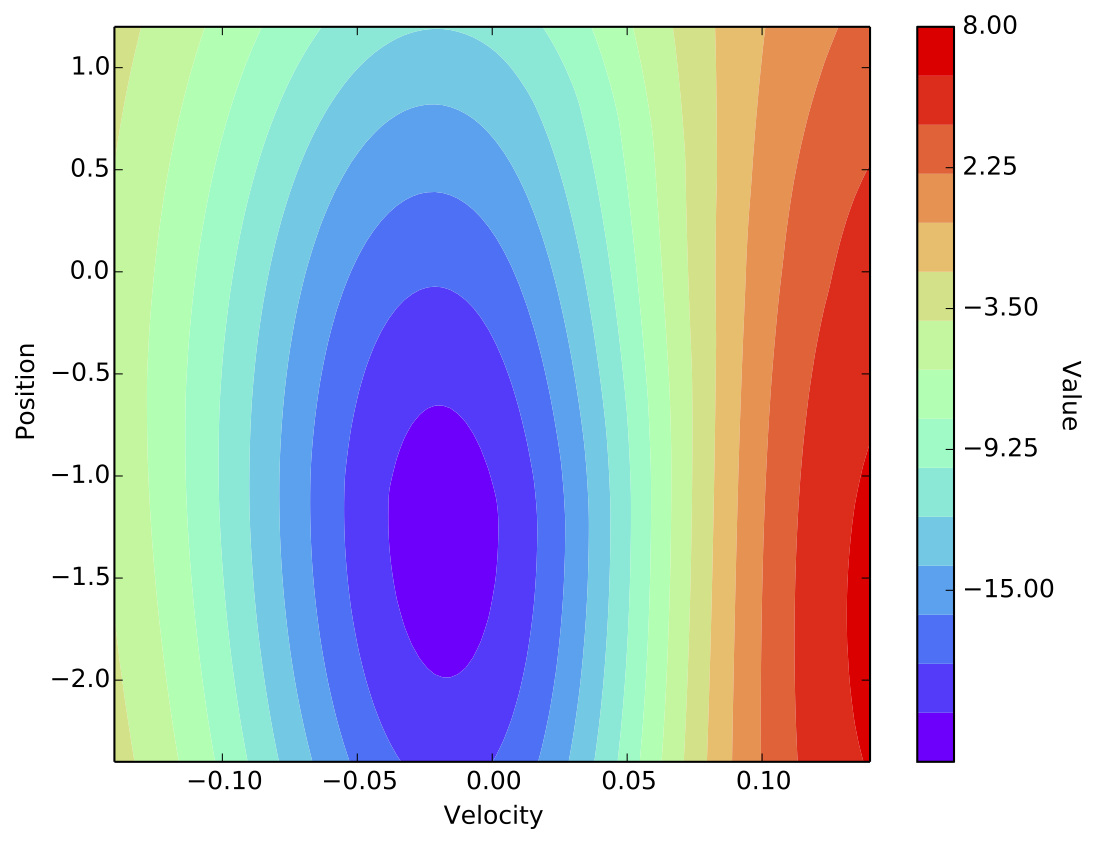}
		\includegraphics[width=0.3\textwidth]{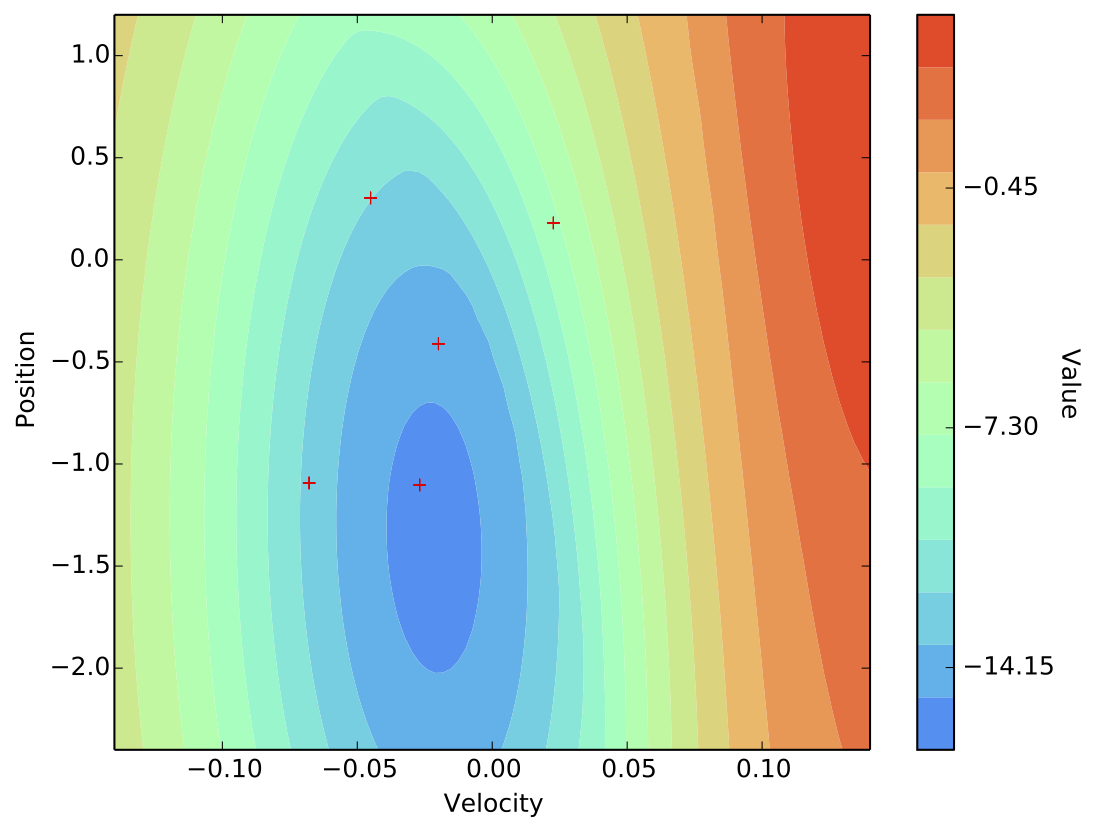}
		\includegraphics[width=0.3\textwidth]{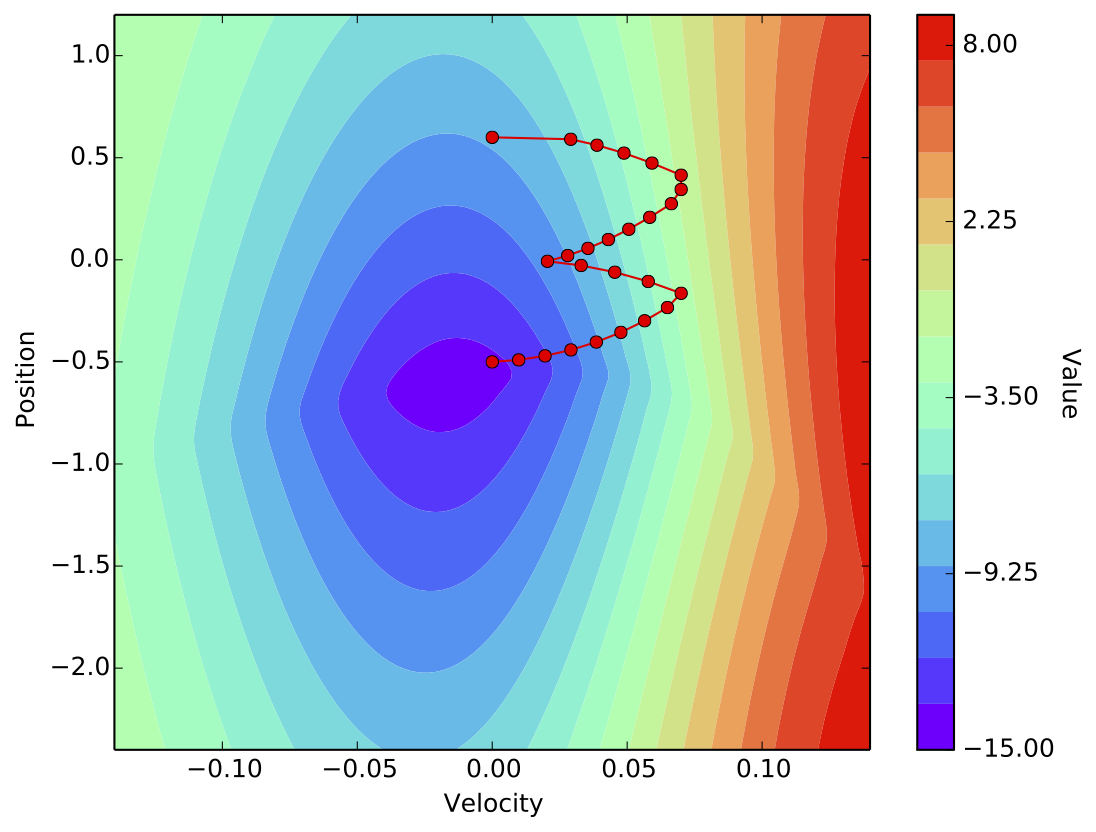}
	\caption{\textbf{\textsc{spgp-sarsa} replicates value functions with a small active set:} Using the same covariance parameters, we plot value predictions from \textsc{gp-sarsa} (left) and \textsc{spgp-sarsa} before (center) and after (right) pseudo-input optimization. With only a $5\times 5$ matrix inversion, \textsc{spgp-sarsa} nearly replicates the true value landscape, while \textsc{gp-sarsa} used a $302\times 302$ inversion. The pseudo inputs (red crosses) moved beyond plot boundaries after optimization. A sample trajectory is shown at right.}
	\label{fig:mountain_car_results}
\end{figure*}
\begin{figure*}[]
	\centering
	\begin{subfigure}[t]{0.3\textwidth}
		\includegraphics[width=\columnwidth]{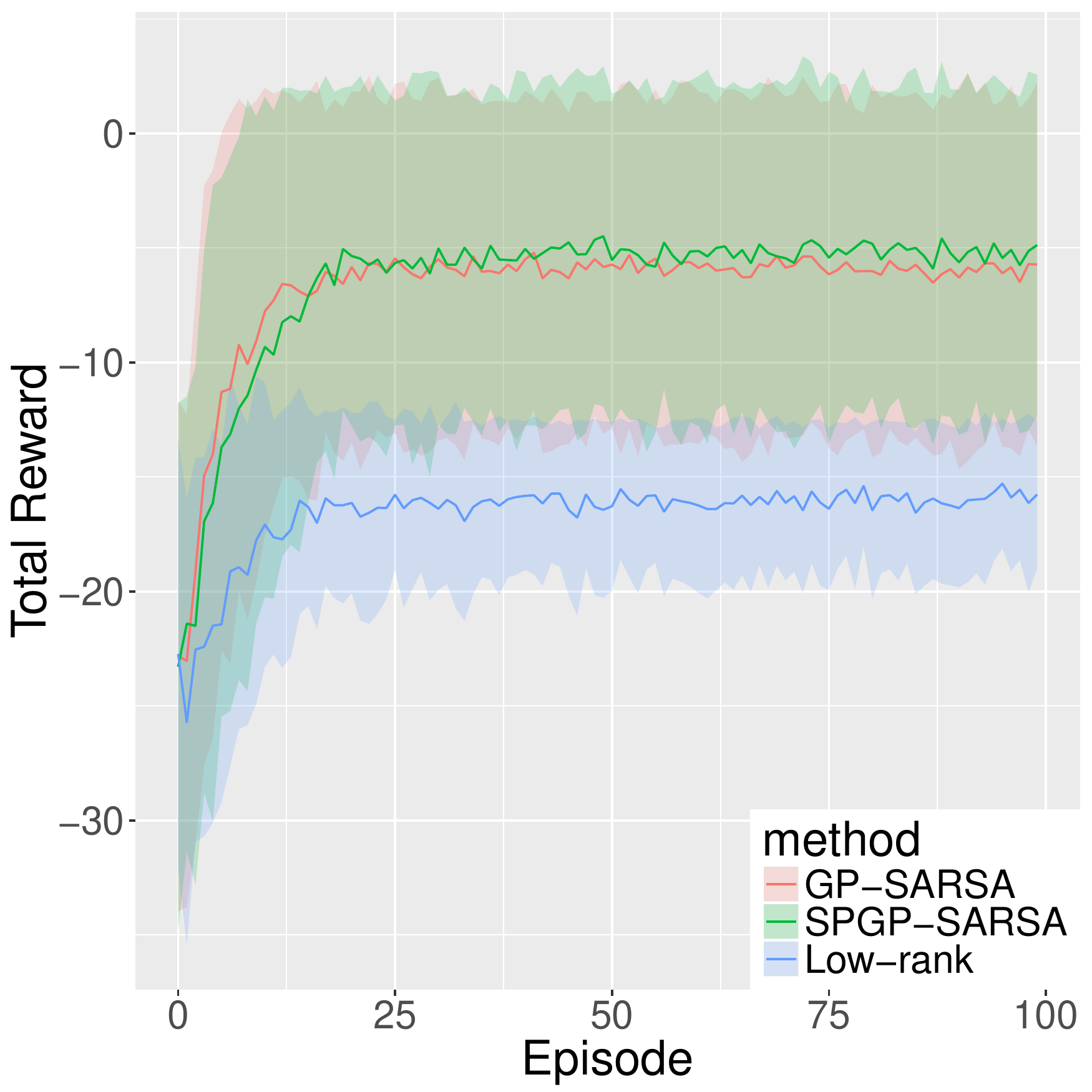}
		\caption{Mountain Car}
	\end{subfigure}
	\begin{subfigure}[t]{0.3\textwidth}
		\includegraphics[width=\columnwidth]{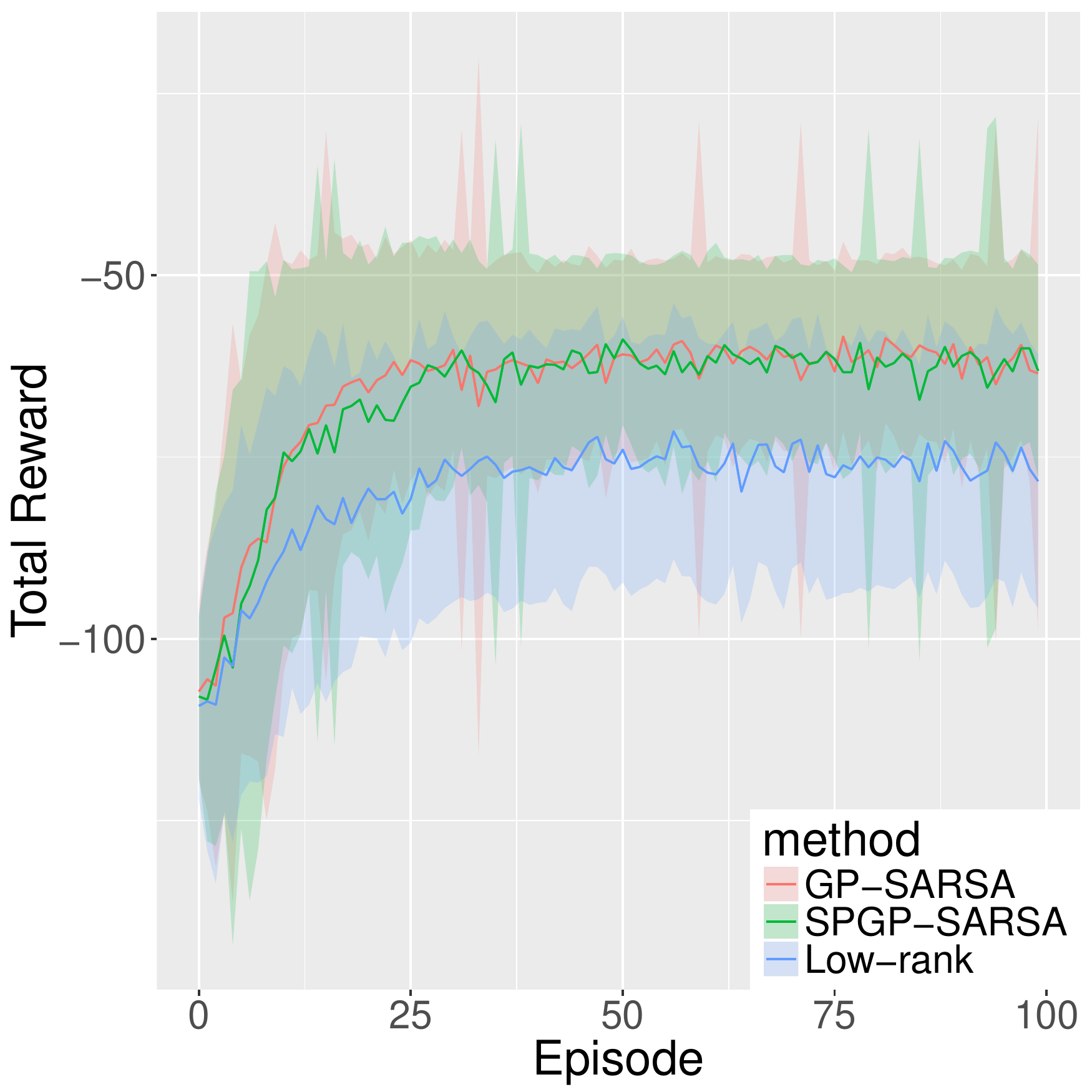}
		\caption{\textsc{usv}}
	\end{subfigure}
	\begin{subfigure}[t]{0.3\textwidth}
		\includegraphics[width=\textwidth]{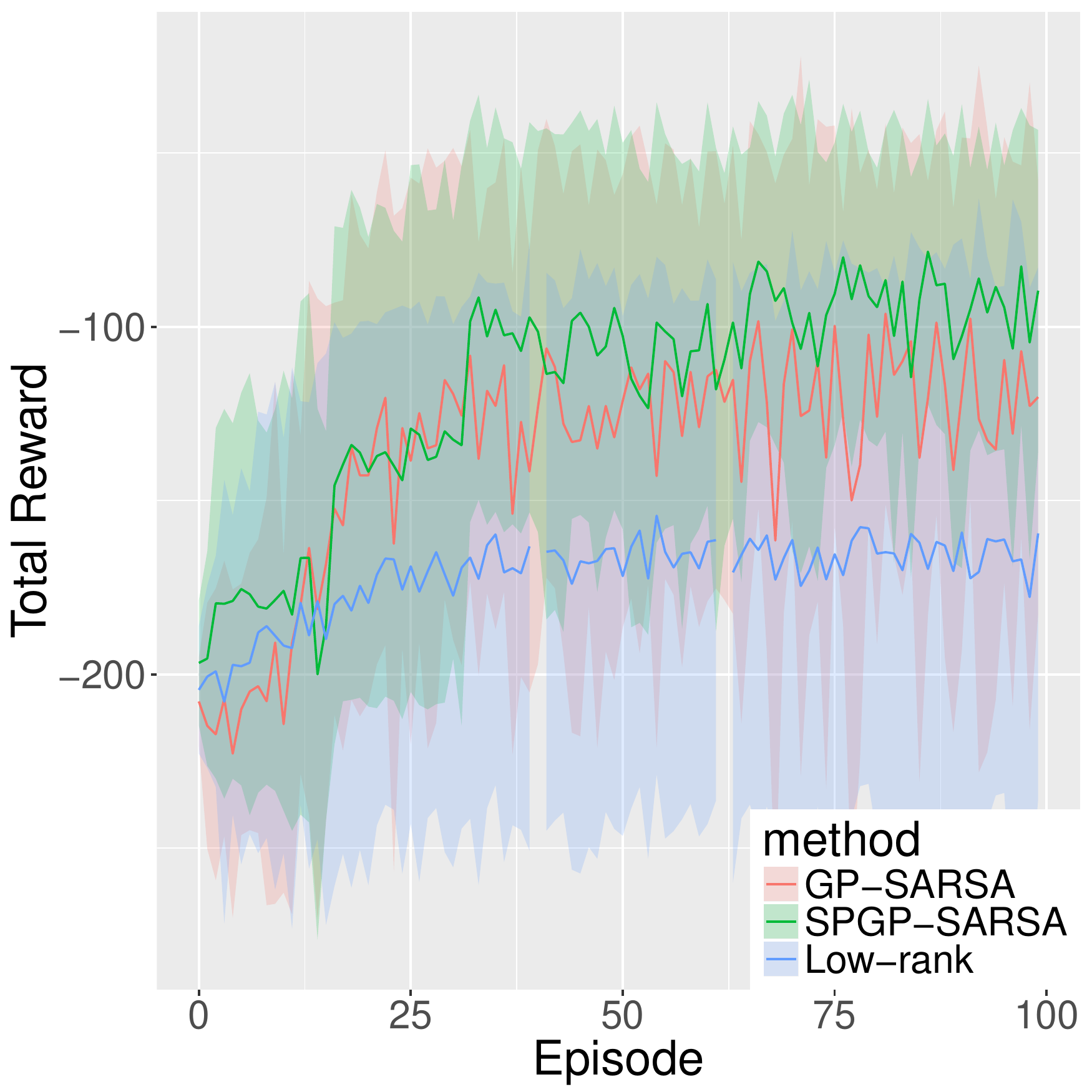}
		\caption{\textsc{uuv}}
	\end{subfigure}
	\caption{\textbf{Learning performance during policy improvement:} We  plot average total reward over 100 episodes, along with one standard deviation. In each case, \textsc{spgp-sarsa} performs as well as \textsc{gp-sarsa}; \textsc{lowrank-sarsa} is the least optimal, and all policies exhibit considerable variance.}
	\label{fig:policy_improvement}
	\vspace{-1em}
\end{figure*}
We presented \textsc{spgp-sarsa} as a method for value estimation and explained how it applies to learning navigation policies. Now, we empirically verify several prior assertions: \textsc{spgp-sarsa} is data efficient; it provides a more flexible and accurate approximation to \textsc{gp-sarsa} than \textsc{lowrank-sarsa}; it is suitable for online applications to marine robots. Evidence to support these claims is provided with several targeted simulation studies and a physical experiment using a BlueROV underwater robot.
 
 

All experiments use the same covariance function: $k(\xbf,\xbf') = \sigma_f\exp{[-0.5(\Delta\xbf^\top\ellbf\Delta \xbf) ]}$, where $\Delta\xbf = (\xbf-\xbf')$, and $\ellbf$ is a diagonal matrix of length scales.  
\subsection{Comparing Approximation Quality}\label{sec:approx_quality}
To facilitate comparison between the approximation quality of \textsc{spgp-sarsa} and \textsc{lowrank-sarsa}, we analyzed a measure of evidence maximization: the ratio of sparse-to-complete log-likelihood, $\Lcal_{\textsc{sparse}}/\Lcal_{\textsc{gptd}}$. We found the low-rank approximation causes sharp changes in the likelihood, and its magnitude often varied around $10^3$. Although this precluded visual comparisons, we are still able to show \textsc{spgp-sarsa} provides a tight approximation. The ratio varies smoothly in relation to different levels of sparsity and improves further as pseudo inputs are optimized (Figure \ref{fig:sparse_tests:likedisp}). 

In a second study, we examined the range of \textsc{lowrank-sarsa}'s adjustability.    In principle, the error threshold $\nu$ can be tuned to any positive number. However, results show the available range can be limited and result in extreme amounts of data retention or rejection  (Figure \ref{fig:sparse_tests:rejstats}).  

As marine robots require learning algorithms that are simultaneously data-efficient and online-viable, it can be problematic if one quality is missing. Retaining an excessive amount of data reduces the computational benefit of the low-rank approximation. Conversely, rejecting too much data is counterproductive when very little arrives. \textsc{spgp-sarsa} offers a good middle ground, because it retains all observations while still achieving fast predictions at any level of sparsity.

\subsection{Simulated Navigation Tasks}
For this experiment, we solve the complete \textsc{rl} problem. Employing Algorithm \ref{alg:policyiter}, we compare performance of each value estimator, $\{$\textsc{gp, spgp, lowrank}$\}$\textsc{-sarsa}, as they inform policy updates on simulated navigation tasks. The purpose is to understand each algorithm's learning performance. 

From the data (Figure \ref{fig:policy_improvement}), it is clear that all algorithms converge quickly: in around fifty episodes. These results are consistent with prior work by Engel \etal \cite{rl:2005:engel_etal:learning_to_control_an_octopus_arm_with_gaussian_process_temporal_difference_methods}, where \textsc{gp-sarsa} was applied to control a high dimensional octopus arm having 88 continuous state variables and 6 action variables. Performance differences between the two approximate methods are due to their approximation quality (Figure \ref{fig:sparse_tests}). As expected, \textsc{spgp-sarsa} is able to learn on par with the exact method, because it can replicate the predictive posterior better than \textsc{lowrank-sarsa}.

In each experiment, robots learn over 100 episodes and select actions with unique $\epsilon$-greedy policies. The number of pseudo inputs, $M$, was selected for a fair comparison. Specifically, after finding a $\nu$ that induced approximately 50\% sparsity, we choose $M$ so both methods converged with active sets of the same size. All pseudo inputs were initialized randomly. 

First, we considered a canonical \textsc{rl} navigation problem, the Mountain Car \cite{rl:1998:sutton_barto:introduction_to_reinforcement_learning}. With limited power, the robot must learn to exploit its dynamics to reach the crest of a hill. The state is given by $\sbf = (s,\dot{s})$, where $s\in[-1.2,0.6]$, and $\dot{s}\in[-0.07,0.07]$. Rewards are $R = \varepsilon - s$, with $\varepsilon\sim\Ncal(0,0.001)$, until the goal is reached, where $R=1$. Episodes start at $\sbf_0 = (-0.5,0.0)$, and the goal is $s_{\text{goal}} = 0.6$. We let $M=5$, $\nu=0.1$, and learning evolve over $50$ transitions. One action, $a\in [-1,1]$ controls the robot's motion (Figure \ref{fig:mountain_car_results}) .


Our second system is a planar Unmanned Surface Vehicle (\textsc{usv}), which has been considered in prior learning experiments \cite{rl:ghavamzadeh_engel_valko-bayesian_policy_gradient_and_actor_critic_algorithm, rl:2017:martin:heteroscedastic}. The robot must navigate within $10\text{m}$ of $\xbf_{\text{goal}}=(50 \text{m},50\text{m})$ using $100$ transitions. States $\sbf =(x,y,\theta,\dot{\theta})$ contain position, $x,y$, heading. $\theta$, and heading rate $\dot{\theta}$. The speed is held constant at $V = 3$ m/s, and the angular rate $\omega\in[-15^\circ/\text{s},15^\circ/\text{s}]$ controls the robot through
\begin{align*}
	x_{t+1} &= x_t +\Delta V\cos{\theta_t}, & y_{t+1} &= y_t +\Delta V\sin{\theta_t},\\
	\theta_{t+1} &= \theta_t +\Delta \dot{\theta}_t, & \dot{\theta}_t &= \dot{\theta}_t +\frac{\Delta}{T}(\omega_t - \dot{\theta}_t).  
\end{align*}
We use time steps of $\Delta = 1.0$s, and a $T = 3$ step time delay for the command $\omega$ to be realized. The delay models resistance of surface currents and actuator limitations. Rewards are assigned with  $R = R_{\text{min}} - ( R_{\text{goal}} - R_{\text{min}} ) \exp( -d / \delta ) + \varepsilon$, where $R_{\text{min}}=-1.0$, $R_{\text{goal}}=10.0$, $\varepsilon$ as before, $d = ||\xbf_{\text{goal}}-\xbf||$, and $\delta =10$. We chose a linear policy, $\omega = K_\omega e_\theta$, where $e_\theta = \arctan{[(50-y)/(50-x)]} - \theta$. $K_\omega$ was updated with a line search, to maximize the first moment of the value posterior. We selected $\nu=0.1$ and $M=50$. 


For the third system we consider a common Unmanned Underwater Vehicle (\textsc{uuv}) design, with differential control. The robot commands forward acceleration, $v$, and turn rate, $\omega$, through port, $a_{\text{port}}$, and starboard, $a_{\text{star}}$, actions. The dynamics are an extension of the \textsc{usv} with the additional dimension, $V_{t+1} = V_t +\Delta v_t$. The policy uses Fourier basis functions, with $e_\theta$ defined as before, and $e_r = || \xbf_{\text{goal}}-\xbf||$:
\begin{align*}
	v&=K_r e_r\cos( e_\theta ), &\omega &= K_r\cos( e_\theta )\sin( e_\theta ) + K_\theta e_\theta. 
	\vspace{-2em}
\end{align*}
These map to actions through $v_t = K_v(a_{\text{port},t}+a_{\text{star},t})$ and $\omega_t =K_w(a_{\text{port},t}- a_{\text{star},t})$, where we let $K_v= K_w=1$. Policy updates select parameters $K_r$ and $K_\theta$ to maximize the value posterior mean - found through a $100\times100$ grid search. Learning occurs over 100 episodes of 200 transitions, with $\nu=5$ and $M=50$.

\subsection{Learning to Navigate with a BlueROV}
\begin{figure*}
	\centering
		\includegraphics[width=0.3\textwidth]{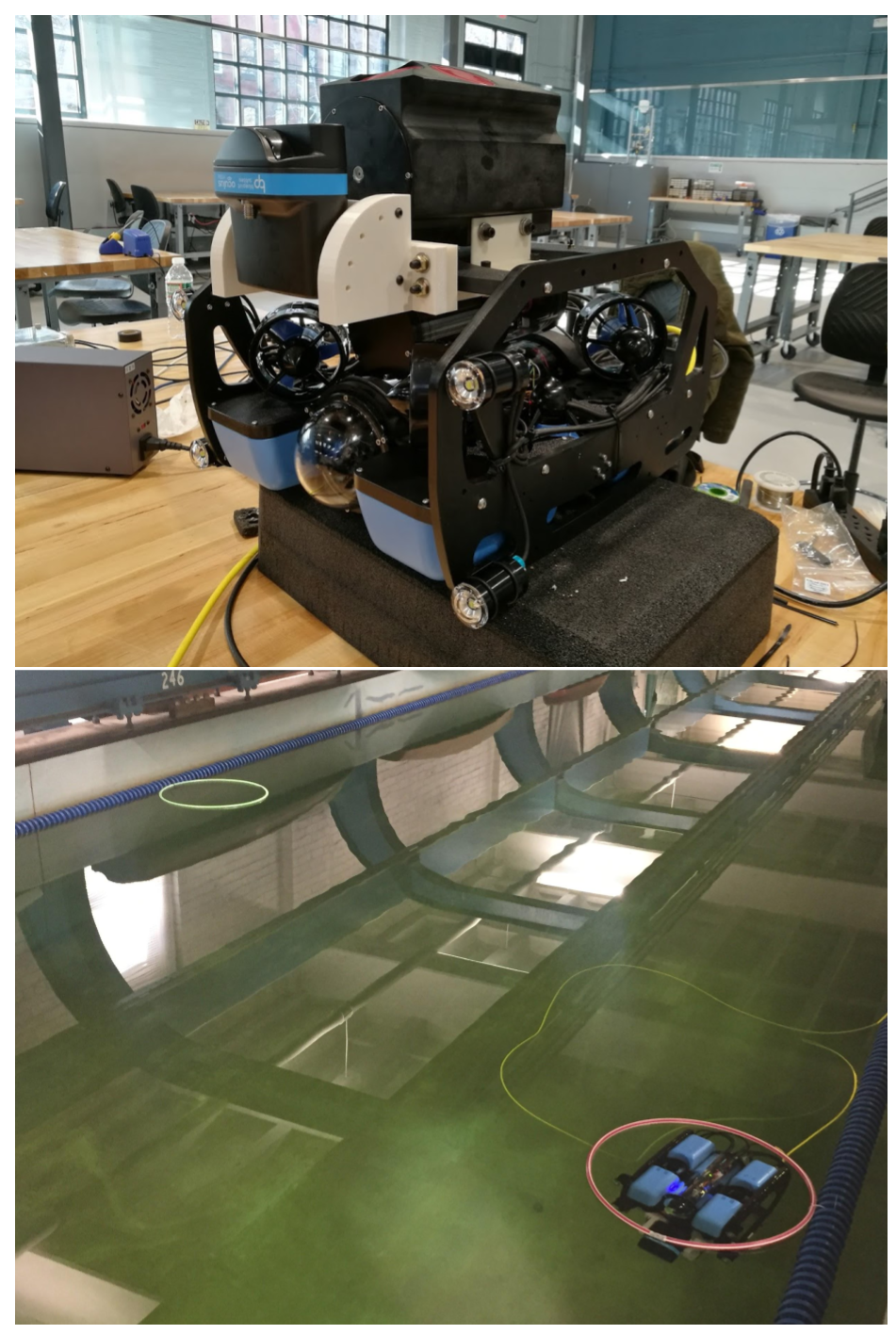}
		\includegraphics[width=0.65\textwidth]{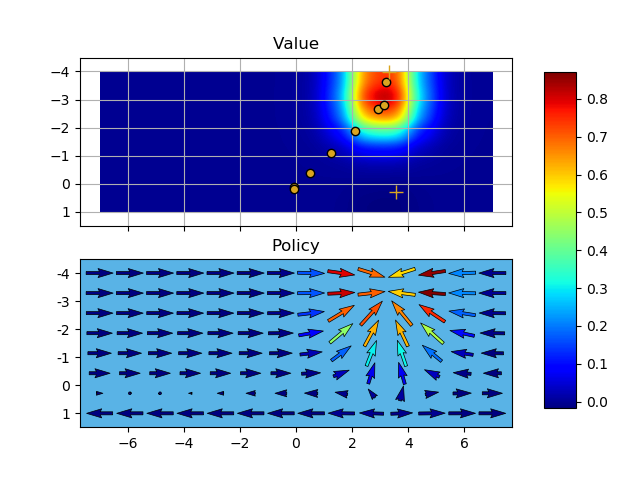}
	\caption{\textbf{Underwater robot navigation:} The robot (top-left) learns to navigate between two rings (bottom-left). With only a demonstration of eight transitions, \textsc{spgp-sarsa} recreates a near-optimal value (top-right) and policy (bottom-right). Robot trajectory and the pseudo inputs are plotted in gold.}
	\label{fig:bluerov_results}
	\vspace{-1.4em}
\end{figure*}

Solving the complete \textsc{rl} problem with a BlueROV (Figure \ref{fig:bluerov_results}) presented a unique set of challenges. States $\sbf=(x,y,\theta)$ derived from a Doppler velocity log, with estimates that drifted on the order of 1m every 1-2 min; this ultimately bounded the number of transitions per episode. While initiating the learning process at depth, disturbances from the data tether introduced uncertainty in the initial position and heading. The robot's speed is also constrained to facilitate accurate localization. To move, the robot can yaw and translate back and forth for a variable length of time.

With numerous limitations imposed on the learning process, we evaluated what could be learned from a single demonstration of only eight transitions. In practice, learning from demonstrations can reduce trial and error \cite{rl:2009:argall_chernova_veloso_browning:a_survey_of_robot_learning_from_demonstration,rl:2013:kober_bagnell_peters:reinforcement_learning_in_robotics_a_survey}. Results show, even with minimal information, that \textsc{spgp-sarsa} was able to recover a near-optimal value function and policy (Figure \ref{fig:bluerov_results}). The experiment was repeated twenty times and the average policy update time took $0.013\pm7\cdot10^{-4}$s with a 2.8GHz i7 processor. We achieve only a modest improvement in prediction time, since $N=8$, and we use $M=2$ pseudo inputs. Despite this fact, our results confirm \textsc{spgp-sarsa} can support efficient robot learning.  
 



%
%
\vspace{0em}
\section{Conclusion}
This paper presented an algorithm that supports learning navigation policies with very little data. We argued for the use of \textsc{gp-td} algorithms to replace standard Bellman recursion, because non-parametric regression can be more data-efficient for learning value functions. We derived \textsc{spgp-sarsa} as a sparse approximation to \textsc{gp-sarsa} and showed it is more flexible and its predictions are more accurate than the state-of-the-art low-rank method. \textsc{spgp-sarsa} was applied to a physical marine robot and learned a near-optimal value function from a single demonstration. In closing, we believe our results highlight the efficiency of \textsc{gp-td} algorithms and the utility of \textsc{spgp-sarsa} as a marine robot learning method.  

\clearpage
\acknowledgments{We thank the anonymous reviewers for their excellent feedback. This research has been supported in part by the National Science Foundation, grant number IIS-1652064. This work was also supported in part by the U.S. Department of Homeland Security under Cooperative Agreement No. 2014-ST-061-ML0001. The views and conclusions contained in this document are those of the authors and should not be interpreted as necessarily representing the official policies, either expressed or implied, of the U.S. Department of Homeland Security.}


\bibliography{ref}  

\clearpage
\section{Supplement}
\subsection{Model Parameter Optimization}
Most optimization packages require an objective function and its gradient to optimize. In the full paper, we described the object function as the log likelihood of the value posterior. Below, we provide the full details of the corresponding gradient computation. 
\subsection{Objective Gradient}
Let $\Kbf_r =  \Qbf + \sigma^2\Ibf + \Kbf_{ru}\Kbf_{uu}^{-1}\Kbf_{ur}$ and $\xi_j$ be the $j$-th optimization variable. The gradient with respect to $\xi_j$ is
\begin{align}\label{eq:likelihood_deriv}
	\frac{\partial \Lcal}{\partial \xi_j} &= -\frac{1}{2}\tr(\Kbf_r^{-1}\Jbf_r) + \frac{1}{2}\rbf^\top\Kbf_r^{-1}\Jbf_r\Kbf_r^{-1}\rbf.
\end{align}
Here, $\Jbf_r$ is the tangent matrix of $\Kbf_r$ with respect to $\xi_j$. The full equations for computing $\Jbf_r$ are described in the next section. 
\subsection{Likelihood Covariance Tangent Matrix}
Denote $\xi_j$ to be a generic optimization variable. Then the matrix $\Jbf_r$ used in Equation \ref{eq:likelihood_deriv} is given by 
\begin{align*}
	\Jbf_r = \frac{\partial \Qbf}{\partial \xi_j} + \frac{\partial}{\partial \xi_j}\sigma^2\Ibf + \Kbf_{ru}\frac{\partial }{\partial \xi_j}(\Kbf_{uu}^{-1}\Kbf_{ur}) + \Jbf_{ru}(\Kbf_{uu}^{-1}\Kbf_{ur}),\\
	\frac{\partial \Qbf}{\partial \xi_j} = \text{diag}\bigl( \Jbf_{rr} -  \Kbf_{ru}\frac{\partial }{\partial \xi_j}(\Kbf_{uu}^{-1}\Kbf_{ur}) - \Jbf_{ru}(\Kbf_{uu}^{-1}\Kbf_{ur}) \bigr),\\
	\frac{\partial }{\partial \xi_j}(\Kbf_{uu}^{-1}\Kbf_{ur}) = \Kbf_{uu}^{-1}\Jbf_{ur} + \frac{\partial \Kbf_{uu}^{-1}}{\partial \xi_j}\Kbf_{ur},\\
	\frac{\partial \Kbf_{uu}^{-1}}{\partial \xi_j} = -\Kbf_{uu}^{-1}\Jbf_u\Kbf_{uu}^{-1}.
\end{align*}

\end{document}